\documentclass[10pt,twocolumn,letterpaper]{article}

\usepackage{wacv}
\usepackage{times}
\usepackage{epsfig}
\usepackage{graphicx}
\usepackage{amsmath}
\usepackage{amssymb}

\usepackage{multirow}
\usepackage{nicefrac}

\wacvfinalcopy 

\def\wacvPaperID{434} 

\ifwacvfinal\fi
\begin{document}

\title{Plug-and-Play CNN for Crowd Motion Analysis: An Application in Abnormal Event Detection}

\author{Mahdyar Ravanbakhsh$^{ 1}$\thanks{Work done while M. Ravanbakhsh was an intern at DISI, University of Trento.}\\
		{\tt\small mahdyar.ravan@ginevra.dibe.unige.it}
		\and
		Moin Nabi$^{ 2, 3}$\\
		{\tt\small m.nabi@sap.com}
		\and
		Hossein Mousavi$^{ 4, 5}$\\
		{\tt\small hossein.mousavi@iit.it}
		\and
		Enver Sangineto$^{ 2}$\\
		{\tt\small enver.sangineto@unitn.it}
		\and
		Nicu Sebe$^{ 2}$\\
		{\tt\small niculae.sebe@unitn.it}
		\and\\
		$^{1}$ University of Genova, Italy   \hspace{.7cm} $^{2}$ University of Trento, Italy \hspace{.7cm}$^{3}$ SAP SE, Berlin, Germany\\
		$^{4}$Istituto Italiano di Tecnologia, Italy \hspace{1cm} $^{5}$Polytechnique Montr\'eal, Montr\'eal, Canada
	}
\maketitle
\ifwacvfinal\thispagestyle{empty}\fi

\begin{abstract}
  Most of the crowd abnormal event detection methods rely on complex hand-crafted features to represent the crowd motion and appearance. Convolutional Neural Networks (CNN) have shown to be a powerful instrument with excellent representational capacities, which can leverage the need for hand-crafted features. In this paper, we show that keeping track of the changes in the CNN feature across time can be used to effectively detect local anomalies. Specifically, we propose to measure local abnormality by combining semantic information (inherited from existing CNN models) with low-level optical-flow. One of the advantages of this method is that it can be used without the fine-tuning phase. The proposed method is validated on challenging abnormality detection datasets and the results show the superiority of our approach compared with the state-of-the-art methods.
\end{abstract}

\section{Introduction}
\label{sec:intro}
	Crowd analysis gained popularity in the recent years in both academic and industrial communities. This growing trend is also due to the increase of population growth rate and the need of more precise public monitoring systems. In the last few years, the computer vision community has pushed on crowd behavior analysis and has made a lot of progress in crowd abnormality detection \cite{mehran2009abnormal,xu2015learning,lu2013abnormal,emonet2011multi,raghavendra2013anomaly,RabieeHMNMS16,rabiee2016crowd, mousavi2015abnormality,del2016discriminative,chaker2017social,amraee2017anomaly,ResnetCrowd2017}. Most of these methods mainly rely on complex hand-crafted features to represent the crowd motion and appearance. However, the use of hand-crafted features is a clear limitation, as it implies task-specific a priori knowledge which, in case of a complex video-surveillance scene, can be very difficult to define. 
	\begin{figure*}
		\begin{center}
			\includegraphics[width=\linewidth]{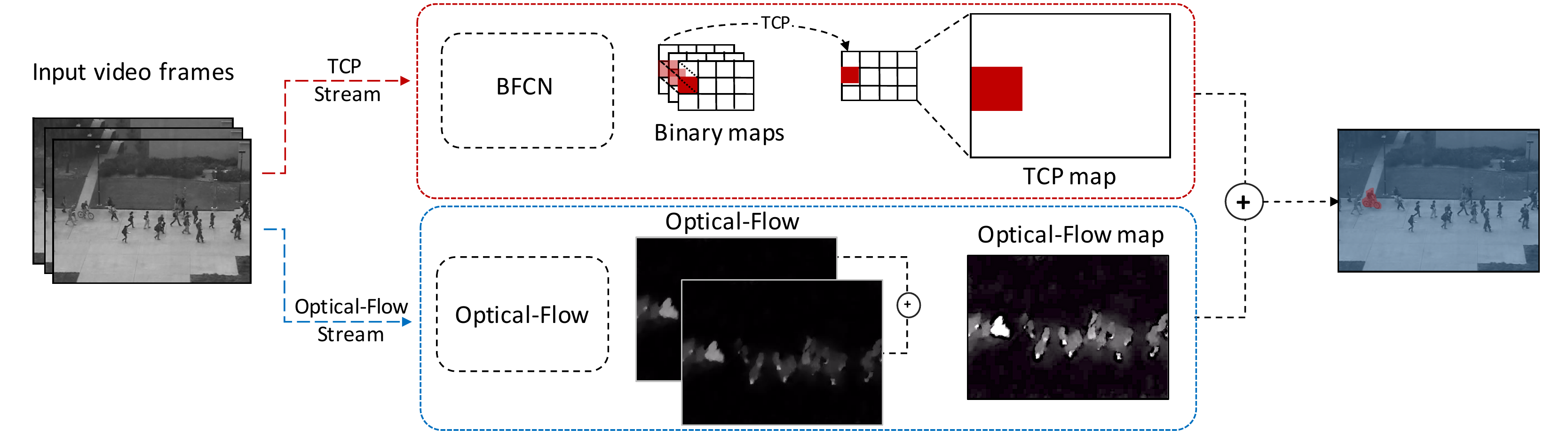}
		\end{center}
		\caption{Overview of the proposed method}
		\label{fig:short}
	\end{figure*}
	Recently, Deep Neural Networks have resurfaced as a powerful tool for learning from big data (e.g., ImageNet~\cite{russakovsky2015imagenet} with 1.2M images), providing models with excellent representational capacities. Specifically, Convolutional Neural Networks (CNNs) have been trained via backpropagation through several layers of convolutional filters. It has been shown that such models are not only able to achieve state-of-the-art performance for the visual recognition tasks in which they were trained, but also the learned representation can be readily applied to other relevant tasks~\cite{razavian2014cnn}. These models perform extremely well in domains with large amounts of training data. With limited training data, however, they are prone to overfitting. This limitation arises often in the abnormal event detection task where scarcity of real-world training examples is a major constraint. Besides the insufficiency of data, the lack of a clear definition of abnormality (\emph{i.e.}, the context-dependent nature of the abnormality) induces subjectivity in annotations. Previous work highlighted the fact that the unsupervised measure-based methods may outperform supervised methods, due to the subjective nature of annotations as well as the small size of training data~\cite{xiong2012energy,sodemann2012review,xiang2008video,mousavi2015crowd}.
	
	Attracted by the capability of CNN to produce a general-purpose semantic representation, in this paper we investigate how to employ CNN features, trained on large-scale image datasets, to be applied to a crowd dataset with few abnormal event instances. This can alleviate the aforementioned problems of supervised methods for abnormality detection, by leveraging the existing CNN models trained for image classification. Besides, training a CNN with images is much cheaper than with videos; therefore, representing a video by means of features learned with static images represents a major saving of computational cost.
	
	The key idea behind our method is to track the changes in the CNN features across time. We show that even very small consecutive patches may have different CNN features, and this difference captures important properties of video motion. To capture the temporal change in CNN features, we cluster them into a set of binary codes each representing a binary pattern {\em(prototype)}. Intuitively, in a given video block consecutive frames should have similar binary patterns unless they undergo a significant motion. We introduced a simple yet effective statistical measure which captures the local variations of appearance in a video block. 
	We show that combining this measure with traditional optical-flow information, provides the complementary information of both appearance and motion patterns.

	\noindent\textbf{Previous Work:} Our method is different from~\cite{mehran2009abnormal,mousavi2015analyzing,Mahadevan.anomaly.2010,cong2011sparse,kim2009observe,saligrama2012video,lu2013abnormal,ResnetCrowd2017}, which focus on learning models on motion and/or appearance features. A key difference compared to these methods is that they employ standard hand-crafted features (\eg, optical-flow, Tracklets, etc.) to model activity patterns, whereas our method proposes using modern deep architectures for this purpose. The advantages of a deep learning framework for anomalous event detection in crowd have been investigated recently in~\cite{xu2015learning, sabokrouFFK16}. Nevertheless, deep networks are data-hungry and need large training datasets. In our work, however, a completely different perspective to abnormality detection is picked out. We specifically propose a measure-based method which allows the integration of semantic information (inherited from existing CNN models) with low-level optical-flow \cite{brox2004high}, with minimum additional training cost. This leads to a more discriminative motion representation while maintaining the method complexity to a manageable level. Most related to our paper is the work by Mousavi \etal~\cite{mousavi2015crowd}, which introduced a similar measure to capture the commotion of a crowd motion for the task of abnormality detection. Instead of capturing the local irregularity of the low-level motion features (e.g., tracklets in~\cite{mousavi2015crowd}) or high-level detectors~\cite{nabi2013temporal}, we propose to represent the crowd motion exploiting the temporal variations of CNN features. This provides the means to jointly employ appearance and motion. Very recently Ravanbakhsh \etal~\cite{ravanbakhsh2015action} proposed a complex feature structure on top of CNN features which can capture the temporal variation in a video for the task of activity recognition. However, to our knowledge this is the first work proposing to employ the existing CNN models for motion representation in crowd analysis.

	\noindent\textbf{Method Overview:} Our method is composed of three steps: 1) Extract CNN-based binary maps from a sequence of input frames, 2) Compute the Temporal CNN Pattern (TCP) measure using the extracted CNN-binary maps 3) The TCP measure fuse with low-level motion features (optical-flow) to find the refined motion segments. 
	
	More specifically, all the frames are input to a Fully Convolutional Network (FCN). Then we propose a binary layer plugged on top of the FCN in order to quantize the high-dimensional feature maps into compact binary patterns. The binary quantization layer is a convolutional layer in which the weights are initialized with an external hashing method. The binary layer produces binary patterns for each patch corresponding to the receptive field of the FCN, called {\em binary map}. The output binary maps preserve the spatial relations in the original frame, which is useful for localization tasks. Then, a histogram is computed over the output binary maps for aggregating binary patterns in a spatio-temporal block. In the next step, an \emph{irregularity} measure is computed over these histograms, called TCP measure. 
	Eventually, all the computed measures over all the video blocks are concatenated, up-sampled to the original frame size, and fused with optical-flow information to localize possible abnormalities. In the rest of this paper we describe each part in detail.\\
	
	\noindent\textbf{Contributions:} Our major contributions: \emph{(i)} We introduce a novel Binary Quantization Layer, \emph{(ii)} We propose a Temporal CNN Pattern measure to represent motion in crowd, \emph{(iii)} The proposed method is tested on the most common abnormality detection datasets and the results show that our approach is comparable with the state-of-the-art methods.

	The rest of the paper is organized as follows: the Binary Quantization Layer is introduced in Sec.~\ref{sec:net}. In Sec.~\ref{sec:tcp} we show the proposed measure, while our feature fusion is shown in Sec.~\ref{sec:fuse}. The experiments and a discussion on the obtained results is presented in Sec.~\ref{sec:exp}. 
	\section{Binary Fully Convolutional Net (BFCN)}
	\label{sec:net}
	In this section, we present the sequential Fully Convolutional Network (FCN) which creates the binary maps for each video frame. The proposed architecture contains two main modules: {\it 1) the convolutional feature maps}, and {\it 2) binary map representations of local features}. In the following, we describe each part in details.

    \subsection{Frame-based Fully Convolutional Network}
	\label{sec:convNet}
	Early layers of convolutions in deep nets present local information about the image, while deeper convolutional layers contain more global information. The last fully connected layers in a typical CNN represent high-level information and usually can be used for classification and recognition tasks. It has been shown that deep net models trained on the ImageNet~\cite{russakovsky2015imagenet} encode semantic information, thus can address a wide range of recognition problems~\cite{razavian2014cnn,donahue2013decaf}. Since, FCNs do not contain fully-connected layers they preserve a relation between the input-image and the final feature-map coordinates. Hence a feature in the output map corresponds to a large receptive field of the input frame. Moreover, FCNs can process input images of different sizes and return feature maps of different sizes as output. In light of the above, this deep network typology is useful to both extract local and global information from an input image and to preserve spatial relations, which is a big advantage for a localization task.
	
\begin{figure*}[t]
		\begin{center}
			\includegraphics[width=\linewidth]{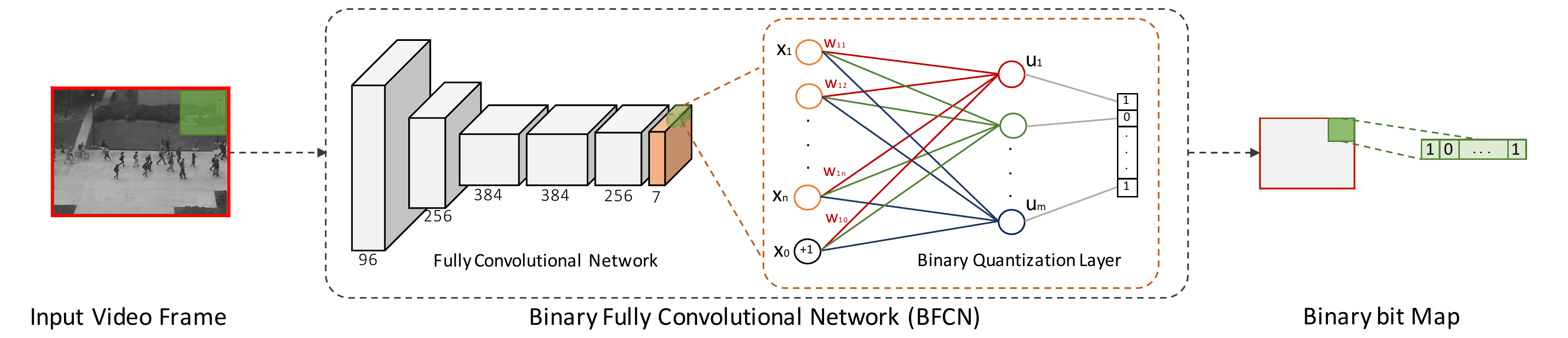}
		\end{center}
		\caption{The Fully Convolutional Net with Binary Quantization Layer, is composed of a fully convolutional neural network followed by a Binary Quantization Layer (BQL). The BQL (shown in orange) is used to quantize the $pool5$ feature maps into 7-bit binary maps.}
		\label{fig:net}
	\end{figure*}

	\noindent\textbf{Convolutional Feature maps:}
	To tackle the gap between the raw-pixel representation of an image and its high-level information we choose the output of the last convolutional layer to extract feature maps. These components provide global information about the objects in the scene. To extract convolutional feature maps, we used a pre-trained AlexNet~\cite{alexnet} model. AlexNet contains 5 convolutional layers and two fully connected layers. In order to obtain spatially localizable feature maps, we feed the output feature maps of the last convolutional layer into our binary quantization layer. Fig.~\ref{fig:net} illustrates the layout of our network.
	\subsection{Binary Quantization Layer (BQL):}
	\label{sec:binary}
	In order to generate a joint model for image segments there is a need to cluster feature components. Clustering the extracted high-dimensional feature maps comes with a high computational cost. The other problem with clustering is the need to know a priori the number of cluster centres. One possible approach to avoid extensive computational costs and to obtain reasonable efficiency is clustering high-dimensional features with a hashing technique to generate small binary codes~\cite{itq}. A 24-bits binary code can address $2^{24}$ cluster centres, which is very difficult to be handled by common clustering methods. Moreover, this binary map can be simply represented as a 3-channels RGB image. Dealing with binary codes comes with a lower computational cost and a higher efficiency compared with other clustering methods. The other advantage of using a hashing technique in comparison with clustering is the ability of embedding the pre-trained hash function/weights as a layer inside the network.
	
	Encoding feature maps to binary codes is done using Iterative Quantization Hashing (ITQ)~\cite{itq}, which is a hashing method for binary code unsupervised learning. Training ITQ is the only training cost in the proposed method, which is done only once on a subset of the train data. ITQ projects each high-dimensional feature vector into a binary space. We use the hashing weights, which are learned using ITQ, to build our proposed Binary Encoding Layer (denoted by $hconv6$). 
	Specifically, inspired by~\cite{2016icip, long2015fully} we implement this layer as a set of convolutional filters (shown in different colors in Fig. \ref{fig:net}), followed by a sigmoid activation function. The number of these filters is equal to the size of the binary code and the weights are pre-computed through ITQ. Finally, the binarization step has been done externally by thresholding the output of the sigmoid function.
	
	Specifically, if $X=\{x_{1},x_{2},...,x_{n}\}$ is a feature vector represented in $pool5$, the output of $hconv6$ is defined by $hconv6(X) = XW_{i}$, where $W_{i}=\{w_{i1},w_{i2},...,w_{in}\}$ are the weights for the $i^{th}$ neuron. The non-linearity is provided by a sigmoid function $\upsilon= \sigma(hconv6(X))$, and the threshold function is defined by:
	\begin{equation}
	\label{eq:activation}
	g(\upsilon) = \Big\{
	\begin{tabular}{cc}
	0, & $\upsilon \le 0.5$\\
	1, & $\upsilon > 0.5$
	\end{tabular}
	\end{equation}
	Eventually, for any given frame our network returns a binary bitmap, which can be used for localization tasks. 
	
	Such a binary quantization layer can be plugged into the net as a pre-trained module, with the possibility of fine-tuning with back-propagation in an end-to-end fashion.However, in our abnormality task, due to the lack of data, fine-tuning is difficult and can be harmful because of possible overfitting, so all our experiments are obtained without fine-tuning.
	
	\noindent\textbf{Sequential BFCN:} Let $\textbf{v} = \{f_{t}\}_{t=1}^{T}$ be an input video, where $f_{t}$ is the $t$-th frame of the video, and $T$ is the number of frames. The frames of the video are fed into the network sequentially. The output for a single frame $f_{t} \in \textbf{v}$, is an encoded binary bit map (prototype), denoted as $m_{t}$. All the binary maps are stacked and provided as the final representation of a given video, \emph{i.e.}, $\textbf{M} = \{m_{t}\}_{t=1}^{T}$.
	\section{Temporal CNN Pattern (TCP)}
	\label{sec:tcp}
	In this section we describe our proposed method to measure abnormality in a video sequence.
	
	\noindent\textbf{Overlapped Video Blocks:}
	The first step is extracting video blocks from the video clip. As mentioned in Sec.~\ref{sec:net}, for a given video frame $f_{t}$ the output of our FCN is a binary bit map denoted by $m_{t}$, which is smaller in size than the original image. In the binary bit map representation, each pixel describes a corresponding region in the original frame. This feature map partitions the video frame into a certain number of regions, which we called $patch$ denoted by $p_{t}^{i}$ where $t$ is the frame number and $i$ is the $i$-th patch in the frame. $b_{t}^{i}$ is a set of corresponding patches along consecutive frames. The length of a video blocks is fixed, and the middle patch of the video block is selected to indicate the entire video block. If the length of a video block $b_{t}^{i}$ is $L+1$, it starts \nicefrac{L}{2} frames before the frame $t$ and ends \nicefrac{L}{2} frames after that, namely $\{b_{t}^{i}\}=\{p_{l}^{i}\}_{l=t-L/2}^{t+L/2}$. To capture more fine information, the video block $b_{t}^{i}$ has $n$ frames overlapped with the next video block $b_{t+1}^{i}$.
		\begin{table*}
		\begin{center}
			\begin{tabular}[width=\textwidth]{|p{6.5cm}|cc|cc|cc|}
				\hline
				\multirow{2}{*}{Method} &\multicolumn{2}{|c|}{Ped1 (frame level)} &\multicolumn{2}{|c|}{Ped1 (pixel level)} &\multicolumn{2}{|c|}{Ped2 (frame level)}\\
				& ERR & AUC & ERR & AUC & ERR & AUC\\
				\hline\hline                                    
				MPPCA~\cite{kim2009observe} & 								40\%    & 	    59.0\%  & 		81\%    &   20.5\%  &       30\%    & 69.3\%    \\
				Social force(SF)~\cite{mehran2009abnormal} & 				31\%    & 	    67.5\%  & 		79\%    &   19.7\%  & 		42\%    & 55.6\%    \\
				SF+MPPCA~\cite{Mahadevan.anomaly.2010} & 			        32\%    & 	    68.8\%  & 		71\%    &   21.3\%  & 		36\%    & 61.3\%    \\
				SR~\cite{cong2011sparse} & 		                            19\%    &       ---     & 		54\%    &   45.3\%  & 		---     & ---       \\
				MDT~\cite{Mahadevan.anomaly.2010} & 		                25\%    &       81.8\%  & 		58\%    &   44.1\%  & 		25\%    & 82.9\%    \\
				LSA~\cite{saligrama2012video} & 	                        16\%    & 	    92.7\%  & 		---     & 	---     & 		---     & ---       \\
				Detection at 150fps~\cite{lu2013abnormal} & 				15\%    & 	    91.8\%  & 		43\%    &   63.8\%  & 		---     & ---       \\
				AMDN (early fusion)~\cite{xu2015learning} & 				22\%    &       84.9\%  & 	    47.1\%  &   57.8\%  & 		24 \%   & 81.5\%    \\
				AMDN (late fusion)~\cite{xu2015learning} & 				    18\%    & 	    89.1\%  & 		43.6\%  &   62.1\%  & 		19 \%   & 87.3\%    \\
				AMDN (double fusion)~\cite{xu2015learning} & 			    16\%    & 	    92.1\%  & 		40.1\%  &   67.2\%  & 		17 \%   & \textbf{90.8\%}    \\
				SL-HOF+FC~\cite{wang2016anomaly}&                           18\%    & 	    87.45\% & 		\textbf{35\%}    &   64.35\% & 		19\%    & 81.04\%   \\
				Spatiotemporal Autoencoder~\cite{chong2017abnormal}&        12.5\%  & 	    89.9\%  & 		---     &   ---     & 		\textbf{12\%}    & 87.4\%    \\
				Sparse Dictionaries with Saliency~\cite{yu2017abnormal}&    ---     & 	    84.1\%  & 		---     &   ---     & 		---     & 80.9\%    \\
				Compact Feature Sets~\cite{leyva2017video}&                 21.15\% & 	    82\%    & 		39.7\%  &   57\%    & 		19.2\%  & 84\%      \\
                Feng et al.~\cite{feng2017learning}&                        15.1\%  & 	    92.5\%  & 		64.9\%  &   \textbf{69.9\%}  & 		---     & ---       \\
                Turchini et al.~\cite{Turchini2017}&                        24\%    & 	    78.1\%  & 		37\%    &   62.2\%  & 		19\%    & 80.7\%    \\

				\hline
				TCP (Proposed Method) &                                     \textbf{8\%}     &       \textbf{95.7\%}  &       40.8\%  &   64.5\%  &       18\%    & 88.4\%    \\
				\hline
			\end{tabular}
		\end{center}
		\caption{Comparison with state-of-the-art on UCSD dataset: reported ERR (Equal Error Rate) and AUC (Area Under Curve). The values of previous methods are reported from~\cite{xu2015learning}.}
		\label{tbl:results}
	\end{table*}
	
	\noindent\textbf{Histogram of Binary Codes:}
	A video block is represented with a set of binary codes (prototypes), which encode the similarity in appearance. We believe that observing the changes of prototypes over a time interval  is a clue to discover motion patterns. Toward this purpose, for each video block $b_{t}^{i}$ a histogram $h_{t}^{i}$ is computed to represent the distribution of prototypes in the video block.
	
	\noindent\textbf{TCP Measure:}
	Similarly to the commotion measure~\cite{mousavi2015crowd}, to obtain the TCP measure for a given video block $b_{t}^{i}$, the \emph{irregularity} of histogram $h_{t}^{i}$ is computed. This is done by considering the fact that, if there is no difference in the appearance, then there is no change in descriptor features and consequently there is no change in the prototype representation. When the pattern of binary bits changes, it means that different appearances are observed in the video block and this information is used to capture motion patterns. The \emph{irregularity} of the histogram is defined as the non-uniformity of the distribution in the video block. A uniform distribution of a histogram shows the presence of several visual patterns in a video block. The higher diversity of the prototypes on a video block leads to a low \emph{irregularity} of the histogram. More uniform histograms increase the chance of abnormality. Such \emph{irregularity} in appearance along the video blocks either is generated by noise or is the source of an anomaly. We took advantage of this fact to present our TCP measure.
	
	The TCP measure for each video block $b_{t}^{i}$, is computed by summing over the differences between the prototype samples in $h_{t}^{i}$ and the dominant prototype. The dominant prototype is defined as the most frequent binary code in the video block, which has the maximum value (mode) in the histogram $h_{t}^{i}$ .
	
	Let $H^{n}$ represent the histogram of binary codes of all patches $\{ p_{t}^{i} \}$ in the video block $b_{t}^{i}$ denoted by $\{ H^{n} \}_{n=1}^{N}$, where $N$ is the number of patches in the video block. The aggregated histogram for block $b_{t}^{i}$ compute as $ \mathcal{H}_{t}^{i} = \sum_{n=1}^{N} H^{n}$. The aggregated histogram $ \mathcal{H}_{t}^{i}$ represents the distribution of the appearance binary codes over the video block $b_{t}^{i}$, and the TCP measure compute as follows:
	\begin{equation}
	\label{eq:coapp}
	tcp(b_{t}^{i}) = \sum\limits_{j=1}^{|\mathcal{H}_{t}^{i}|} {|| \mathcal{H}_{t}^{i}(j) - \mathcal{H}_{t}^{i}(j_{max}) ||_{2}^{2}}
	\end{equation}
	where $|.|$ is the number of bins of the histogram, $||.||_{2}$ is the L2-norm, and the dominant appearance index over the video block is denoted by $j_{max}$ ({\em i.e.}, the mode of $\mathcal{H}_{t}^{i}$).
	
	\noindent\textbf{TCP Map:}
	To create a spatial map of the TCP measure $c_{t}$ for any given frame $f_{t}$, the TCP measure is computed for all video blocks $b_{t}^{i}$, and we assign the value of $c_{t}^{i}$ to a patch that is temporally located at the middle of the selected video block. The output $c_{t}$ is a map with the same size as the binary map $m_{t}$ which contains TCP measure values for each patch in the frame $f_{t}$. Finally, the TCP maps are extracted for the entire video footage. We denote the TCP map for frame $f_{t}$ as $c_{t} = \{c_{t}^{i}\}_{i=1}^{I}$, where $I$ is the number of patches in the frame.
	
	\noindent\textbf{Up-sampling TCP Maps:}
	Since the frame will pass through several convolution and pooling layers in the network, the final TCP map is smaller than the original video frame. To localize the exact region there is a need to produce a map of the same size as the input frame. For this reason, the TCP value in the map is assigned to all pixels in the corresponding patch of the frame on the up-sampled TCP map.
			\begin{table}
		\begin{center}
			\begin{tabular}[width=\textwidth]{|p{5cm}|c|}
				\hline
				Method 														& 	AUC \\
				\hline
				\hline
				Optical-Flow~\cite{mehran2009abnormal}	&	0.84 \\
				SFM~\cite{mehran2009abnormal} 				&	0.96\\
				Del Giorno et al.\cite{del2016discriminative} & 0.910 \\
				Marsden et al. \cite{marsden2016holistic} & 0.929\\
				Singh and Mohan~\cite{singh2017graph}& 0.952\\
				Sparse Reconstruction~\cite{cong2011sparse}&0.976\\
				Commotion~\cite{mousavi2015crowd} 		&	\textbf{0.988}\\
				Yu et al.~\cite{yu2017abnormal} 		&	0.972\\
				Cem et al.~\cite{direkoglu2017abnormal} 		&	0.964\\
				\hline
				TCP (proposed method) 							&	\textbf{0.988}\\
				\hline
			\end{tabular}
		\end{center}
		\caption{Results on UMN dataset. The values of previous methods are reported from~\cite{mousavi2015crowd}.}
		\label{tbl:auc}
	\end{table}
	\section{Fusion with optical-flow Maps}
	\label{sec:fuse}
	Since the Up-sampled TCP map can only detect the coarse region of abnormality, we propose to fuse optical-flow with the TCP maps in order to have a more accurate localization.
	The optical-flow \cite{brox2004high} is extracted from each two consecutive frames. However the TCP map is computed for each $L$ frames. To be able to fuse the optical-flow with the corresponding extracted TCP map, an aligned optical-flow map is constructed. Suppose that $f_{t}$ and $f_{t+1}$ are two consecutive frames from video $\textbf{v} = \{f_{t}\}_{t=1}^{T}$, optical-flow map $of_{t}$, with the same resolution of an input frame, represents the optical-flow values transition between the two frames. optical-flow values are extracted for entire video footage $\textbf{v}$ and stacked as optical-flow sequences $\{d_{t}\}_{t=1}^{T-1}$. Finally, similar to the overlapped video block extraction protocol, overlapped optical-flow maps are computed. If the length of a video block $p_{t}^{i}$ is $L+1$, then the corresponding optical-flow map $d_{t}^{i}$ is the sum of all optical-flow values over the corresponding $i$-th region as $d_{t}^{i} = \sum_{l=1}^{L} {d_{t}^{i}(l)}$. The optical-flow map for entire frame $f_{t}$ is described as $d_{t} = \{d_{t}^{i}\}_{i=1}^{I}$.
	
	\begin{figure*}
		\begin{center}
			\begin{tabular}{c c}
				\includegraphics[width=0.5\linewidth]{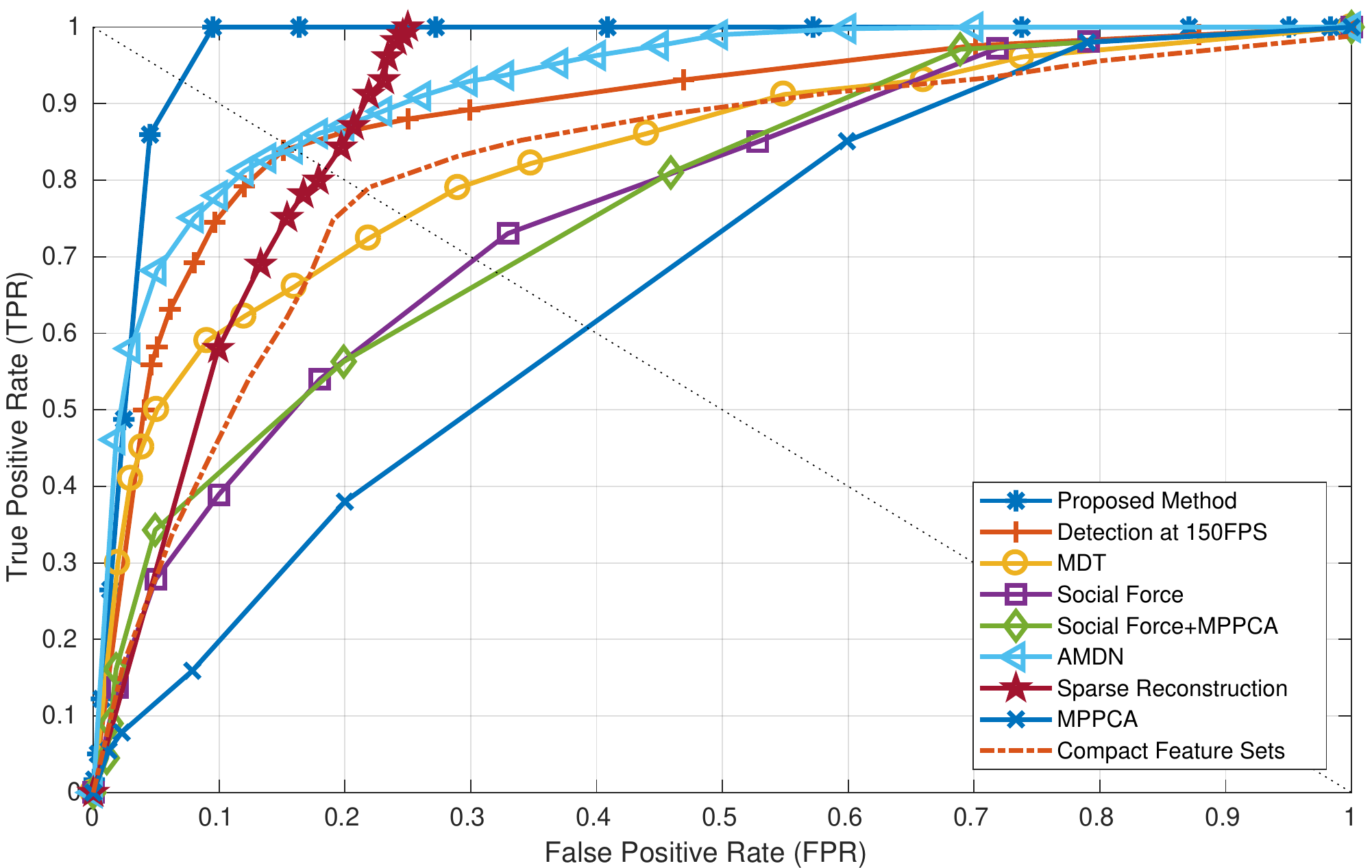} & \includegraphics[width=.5\linewidth]{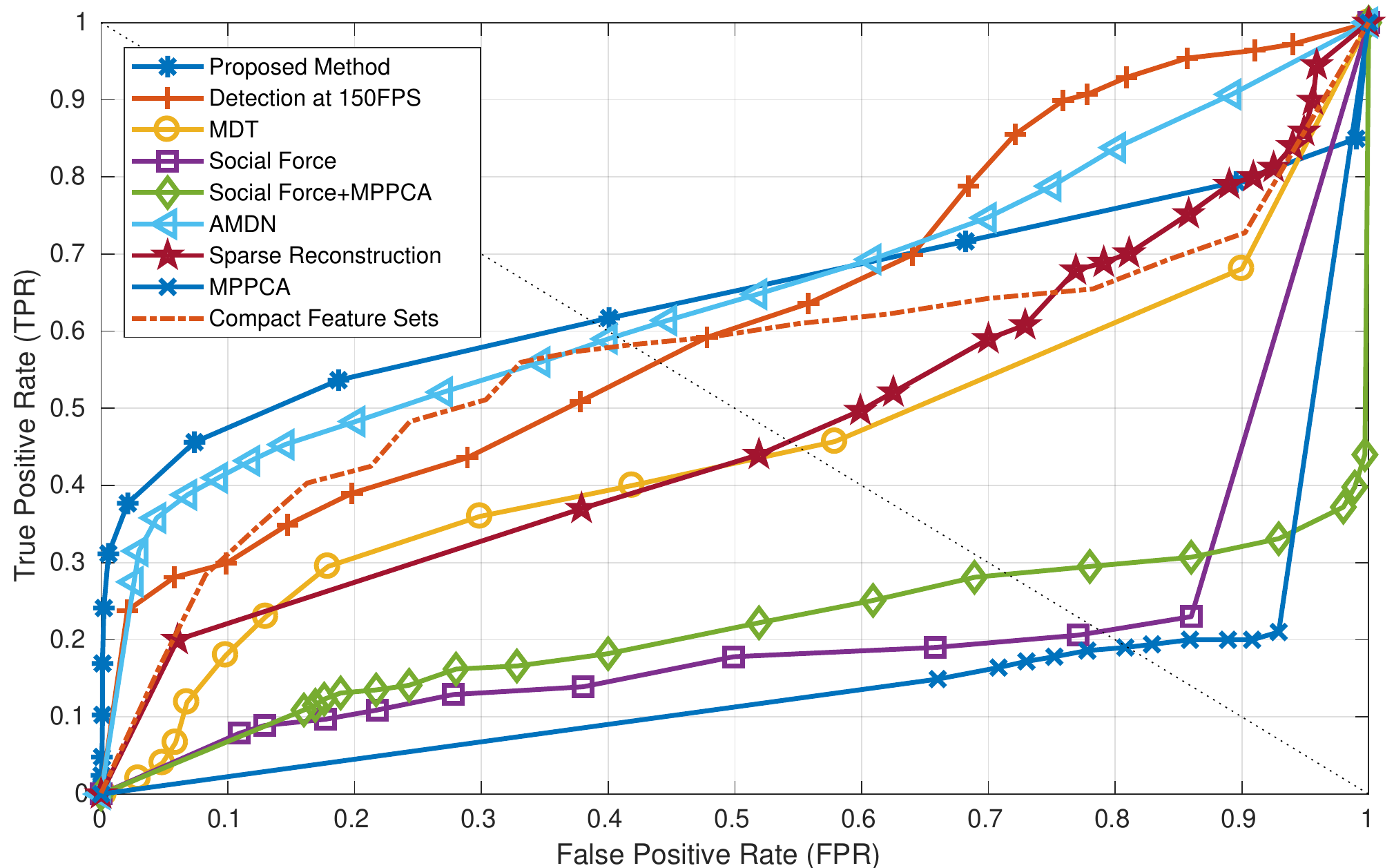} \\ 
				{\scriptsize(a) ROC curve of frame level (anomaly detection)} & {\scriptsize(b) ROC curve of pixel level (anomaly localization)}\\
			\end{tabular}	
		\end{center}
		\caption{Frame level and Pixel level comparison ROC curves of Ped1 (UCSD dataset).}
		\label{fig:rocfrm}
	\end{figure*}
	
	\noindent\textbf{Feature Fusion:}
	The extracted optical-flow maps and the computed TCP maps for each video frame are fused together with importance factors $\alpha$ and $\beta$ to create motion segment map: $mseg_{t} = \alpha d_{t} + \beta c_{t} \;, \; mseg = \{mseg_{t}\}_{t=1}^{T}$, where, $\{mseg\}$ is the extracted motion segments along the entire video $\textbf{v}$. The importance factors indicates the influence of each fused map in the final segment motion map, we simply select $0.5$ for both $\alpha$ and $\beta$.
	
	\section{Experimental Results}
	\label{sec:exp}
	In this section, we evaluate our method over two well-known crowd abnormality datasets and compare our results with state of the art. The evaluation has been performed with both a {\em pixel-level} and a {\em frame-level} protocol, under standard setup. The rest of this section is dedicated to describing the evaluation datasets, the experimental setup and the reporting the results quantitatively and qualitatively.
	
	\noindent\textbf{Datasets and Experimental Setup:}
	In order to evaluate our method two standard datasets: UCSD Anomaly Detection Dataset~\cite{Mahadevan.anomaly.2010} and UMN SocialForce~\cite{mehran2009abnormal}. The \textbf{UCSD dataset} is split into two subsets {\em Ped1} and {\em Ped2}. {\em Ped1} contains 34/16 training/test sequences with frame resolution $238 \times 158$. Video sequences consist of 3,400 abnormal frame samples and 5,500 normal frames. {\em Ped2} includes 16/12 training/test video samples, with about 1,600 abnormal frames and 350 normal samples. This subset is captured from different scenes than {\em Ped1}, and the frames resolution is $360 \times 240$. This dataset is challenging due to different camera view points, low resolution, different types of moving objects across the scene, presence of one or more anomalies in the frames. The \textbf{UMN dataset} contains 11 video sequences in 3 different scenes, and 7700 frames in total. The resolution is $320 \times 240$. All sequences start with a normal scene and end with abnormality section.
	
	In our experiments to initialize the weights of $hconv6$ an ITQ is applied on the train set of UCSD pedestrian dataset with a 7-bits binary code representation, which addresses 128 different appearance classes. Video frames are fed to the BFCN sequentially to extract binary bit maps. All video frames are resized to $460 \times 350$, then BFCN for any given frame returns a binary bit map with resolution $8 \times 5$, which splits the frame into a 40-region grid.
	The length of video block extracted from a binary map is fixed to $L=14$ with 13 frames overlapping. The TCP measure is normalized over the entire video block sequence, then a threshold $th<0.1$ is applied for detecting and subtracting the background region in the video.
	
	Optical-flow feature maps are extracted to fuse with our computed features on the TCP measure maps. The fusion importance factor set to $0.5$ equally for both feature sets. These motion segment maps are used to evaluate the performance of our method on detection and localization of anomalous motions during video frames.
	\subsection{Quantitative Evaluation}
	The evaluation is performed with two different levels: {\em frame level} for anomaly detection, and {\em pixel level} for anomaly localization. We evaluate our method on UCSD abnormality crowd dataset under the original setup~\cite{li2014anomaly}.
		\begin{figure*}
			\begin{center}
				\includegraphics[width=\linewidth]{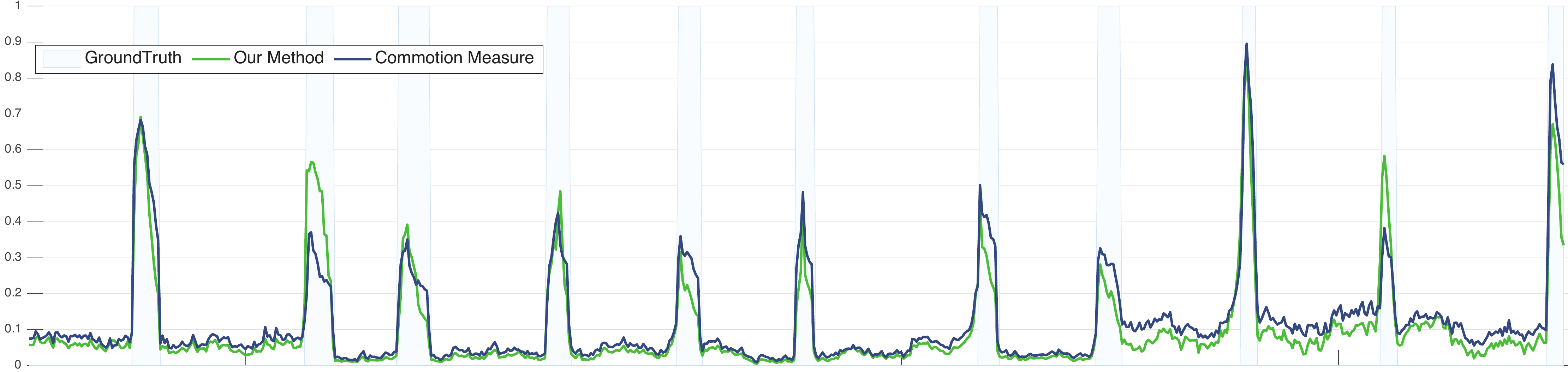}
			\end{center}
			\caption{Frame-level anomaly detection results on UMN dataset: Our method compares to Commotion Measure~\cite{mousavi2015crowd}. The green and blue signals respectively show the computed TCP by our approach and Commotion Measure over frames of 11 video sequences. The light blue bars indicate the ground truth abnormal frames of each sequence. All the sequences start with normal frames and ends with abnormality.}
			\label{fig:umn}
		\end{figure*}
	\noindent\textbf{Frame Level Anomaly Detection:}
	This experiment aims at evaluating the performance of anomaly detection along the video clip. The criterion to detect a frame as abnormal is based on checking if the frame contains at least one abnormal patch. To evaluate the performances the detected frame is compared to ground truth frame label regardless of the location of the anomalous event. The procedure is applied over range of thresholds to build the ROC curve. We compare our method with state-of-the-art in detection performance on UCSD ped1 and ped2 datasets. The result is shown in Table~\ref{tbl:results}, beside the ROC curves on Fig.~\ref{fig:rocfrm}.
	
	The proposed method is also evaluated on UMN dataset. Fig.~\ref{fig:umn} shows the computed TCP for each frame illustrated as ``detection  signal'' (green). We compared TCP with commotion measure (blue). The overall TCP value for a frame is computed from the sum of TCP measures over the patches in a frame and normalized in $[0,1]$ as an abnormality indicator. In Fig.~\ref{fig:umn}, the horizontal axis represents the time($s$), the vertical axis shows the ``abnormality indicator'', and the light blue bars indicate the ground truth labels for abnormal frames.
	
	\noindent\textbf{Pixel Level Anomaly Localization:}
	The goal of the pixel level evaluation is to measure the accuracy of anomalous event localization. Following~\cite{li2014anomaly}, detected abnormal pixels are compared to pixel level groundtruth. A true positive prediction should cover at least 40\% of true abnormal pixels over groundtruth, otherwise counted as a false positive detection.
	Fig.~\ref{fig:rocfrm} shows the ROC curves of the localization accuracy over USDC Ped1 and Ped2. We compare our method with state of the art in accuracy for localization. Result is presented in Table~\ref{tbl:results}.

\begin{figure}
	\begin{center}
		\begin{tabular}{c c}
			\includegraphics[width=\linewidth]{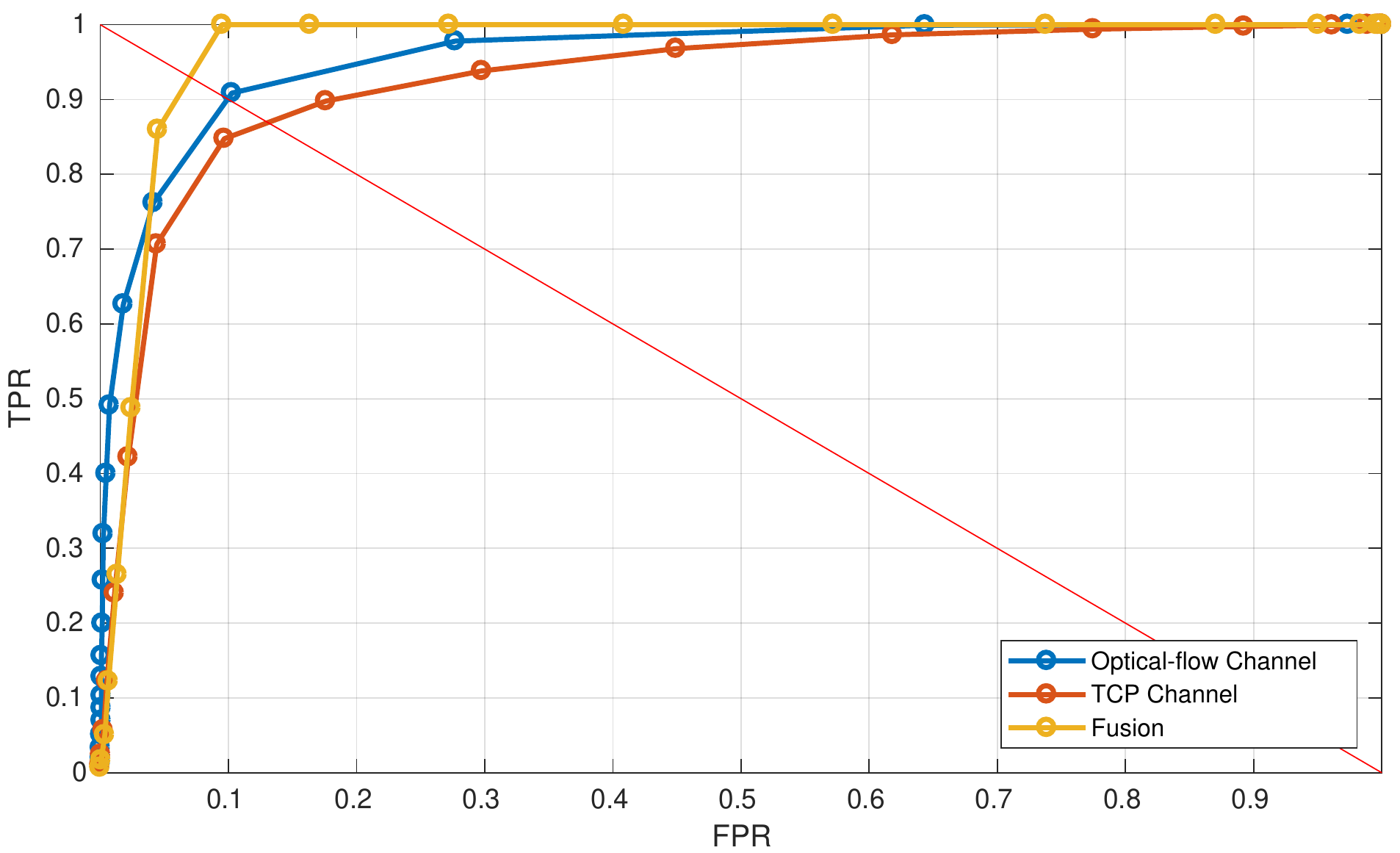} 
		\end{tabular}	
	\end{center}
	\caption{Comparison of different Streams in the proposed method on Ped1 frame-level evaluation.}
	\label{fig:rocstreams}
\end{figure}

\begin{figure*}[t]
		\begin{center}
		(a)
			\includegraphics[width=.235\linewidth]{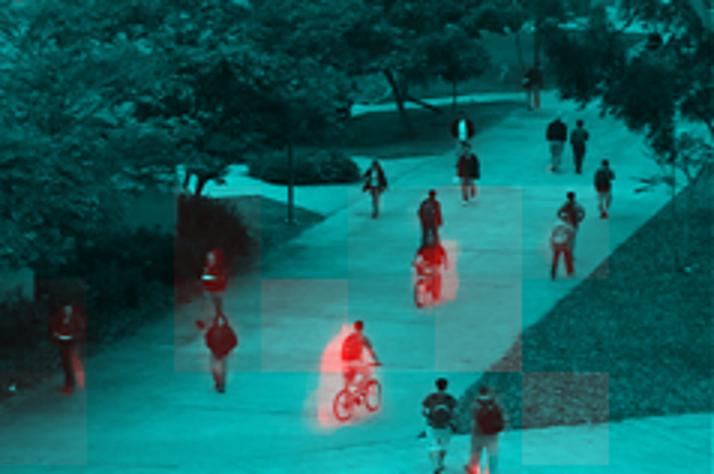}
			\includegraphics[width=.235\linewidth]{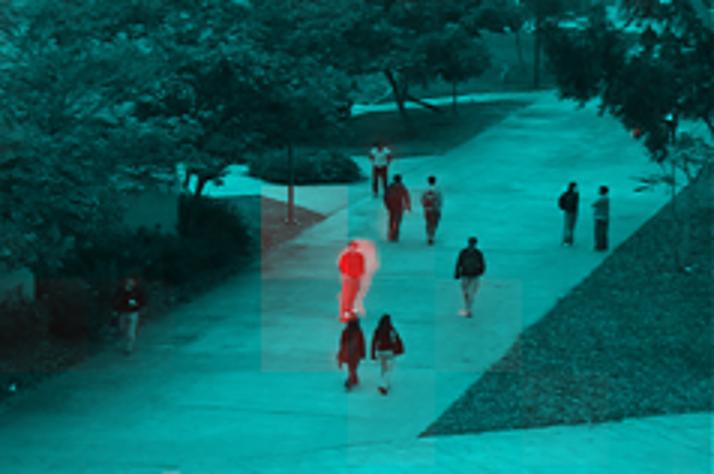}
			\includegraphics[width=.235\linewidth]{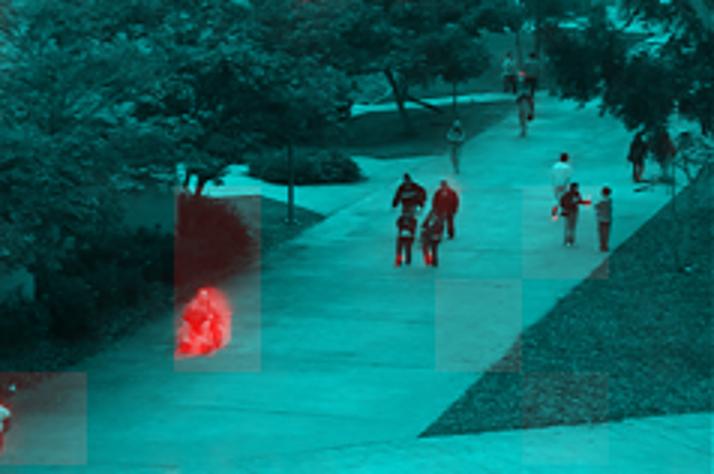}
			\includegraphics[width=.235\linewidth]{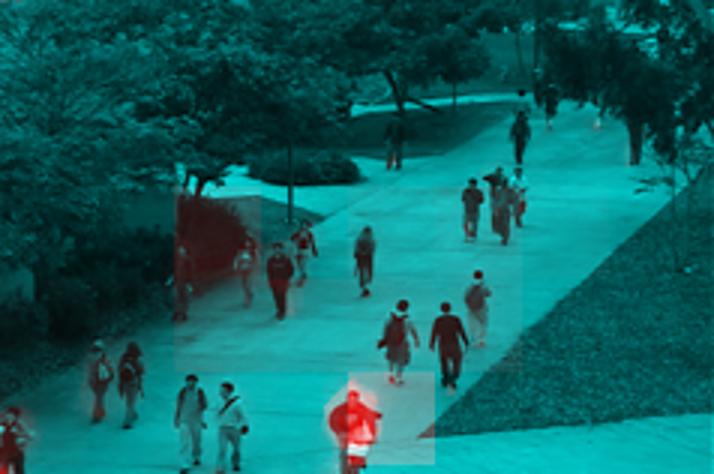}
			
			(b)	
			\includegraphics[width=.235\linewidth]{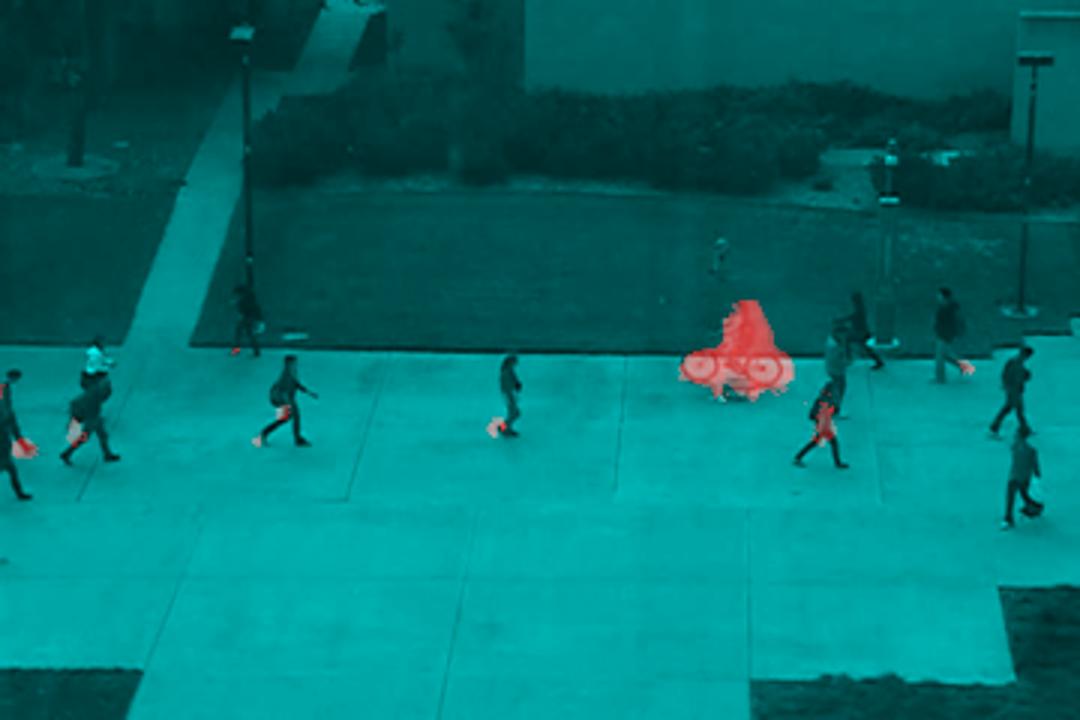}
			\includegraphics[width=.235\linewidth]{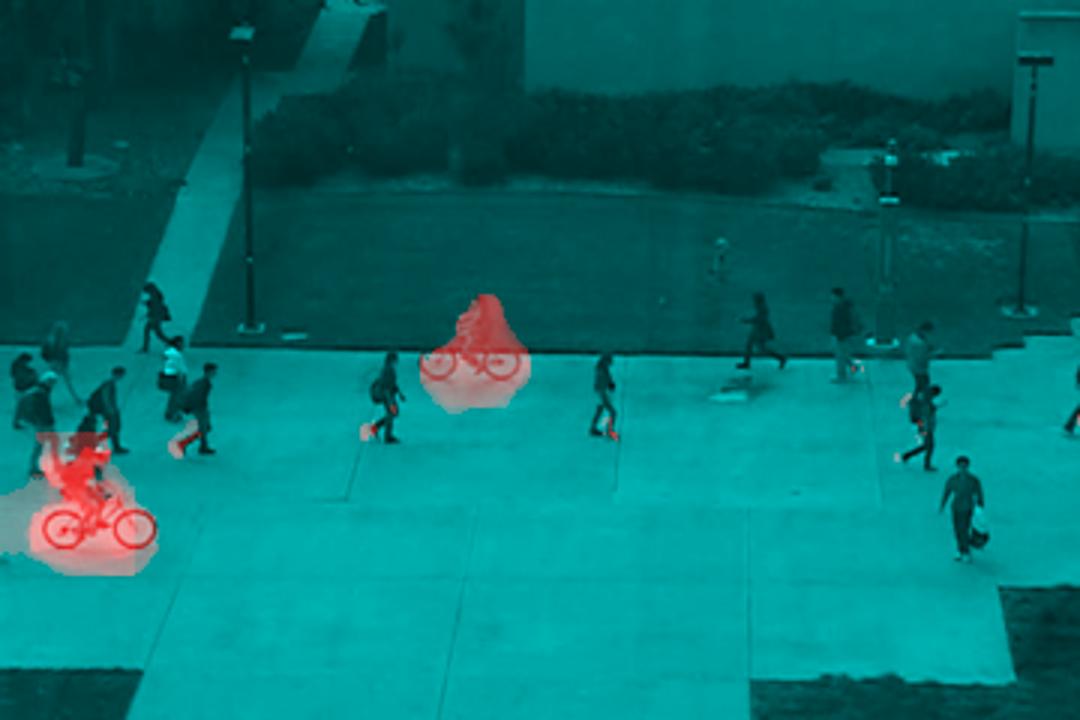}
			\includegraphics[width=.235\linewidth]{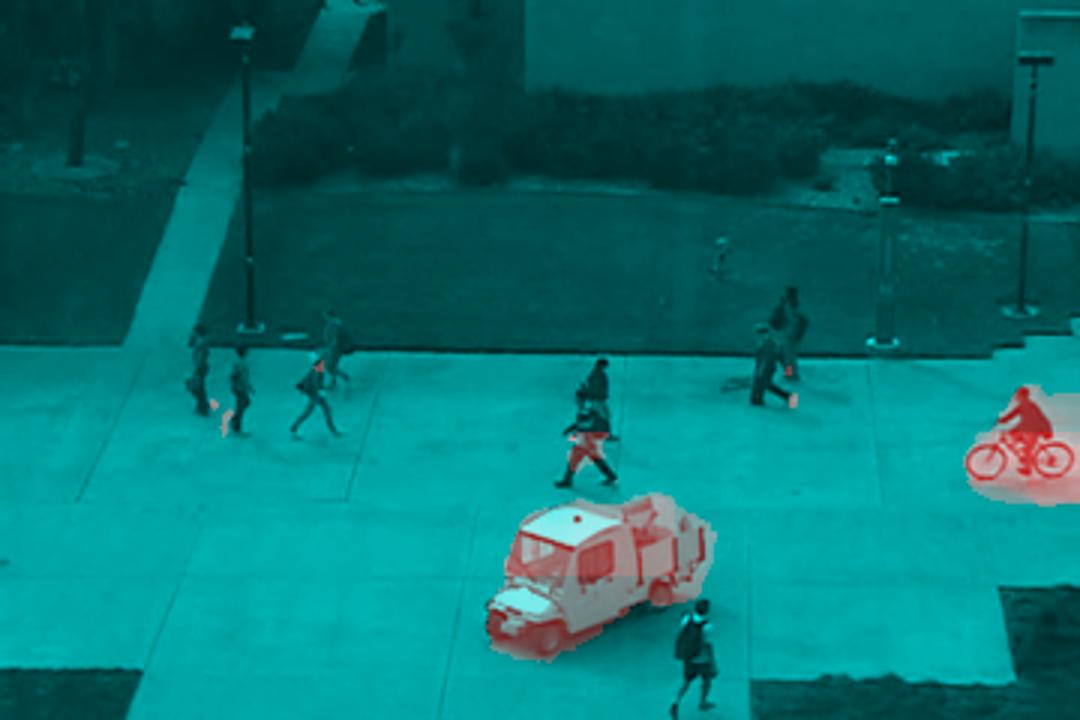}
			\includegraphics[width=.235\linewidth]{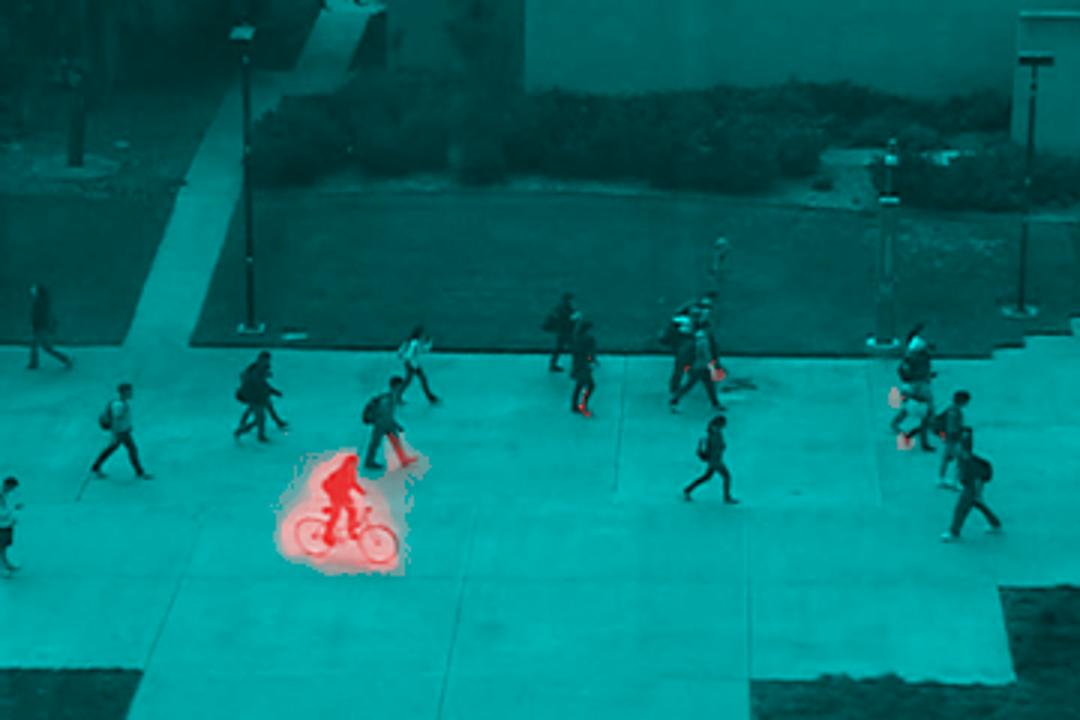}
			
			(c)
			\includegraphics[width=.235\linewidth]{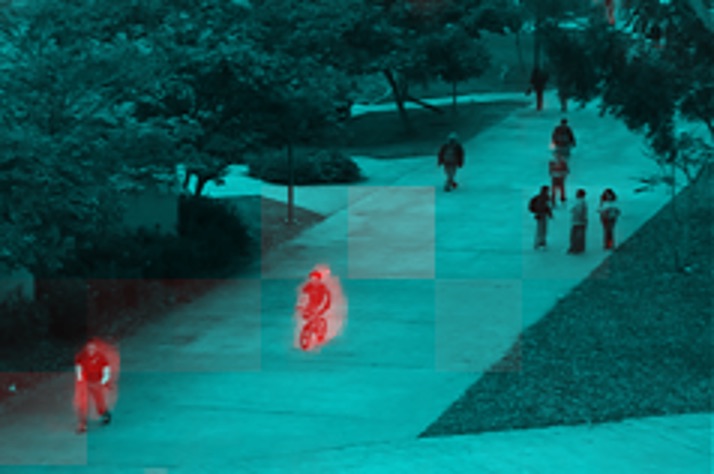}
			\includegraphics[width=.235\linewidth]{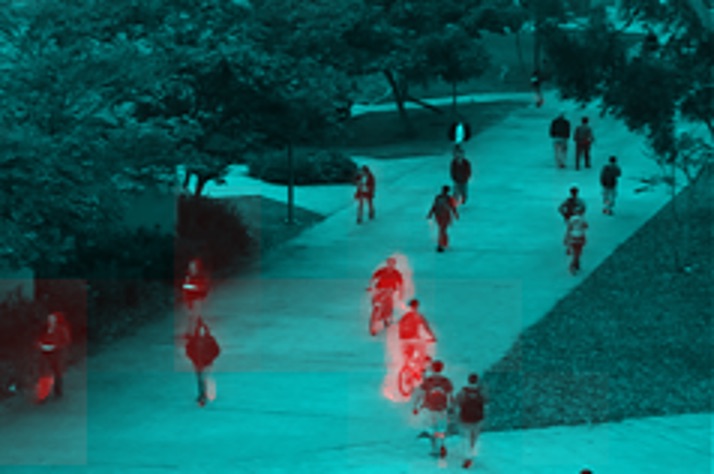}
			\includegraphics[width=.235\linewidth]{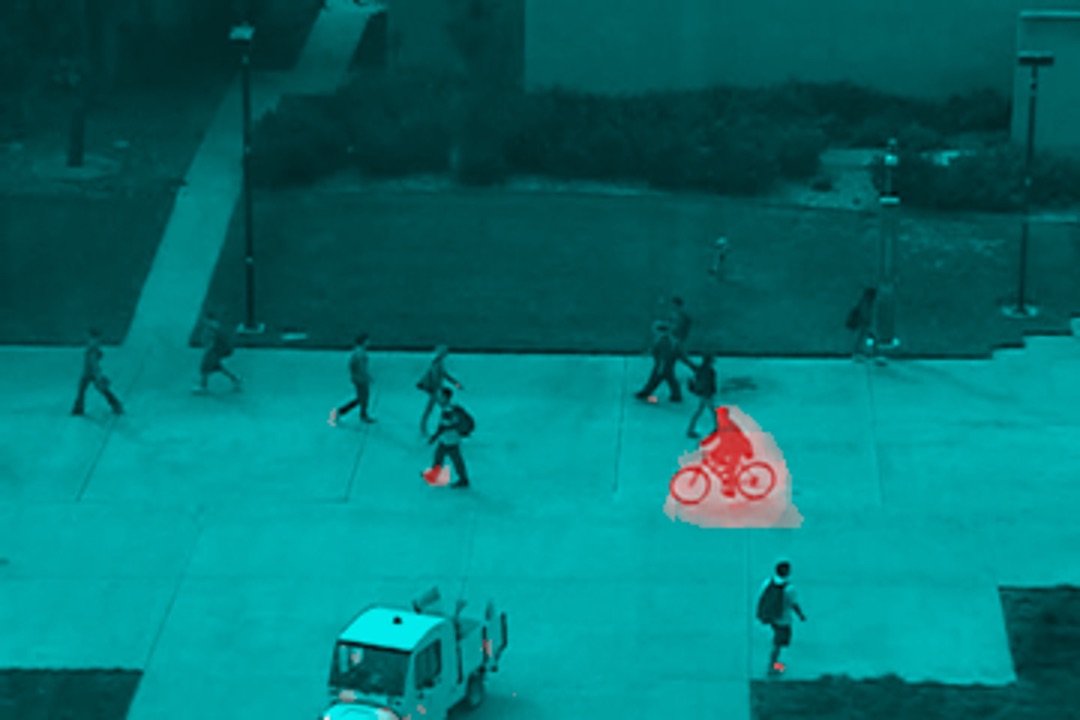}
			\includegraphics[width=.235\linewidth]{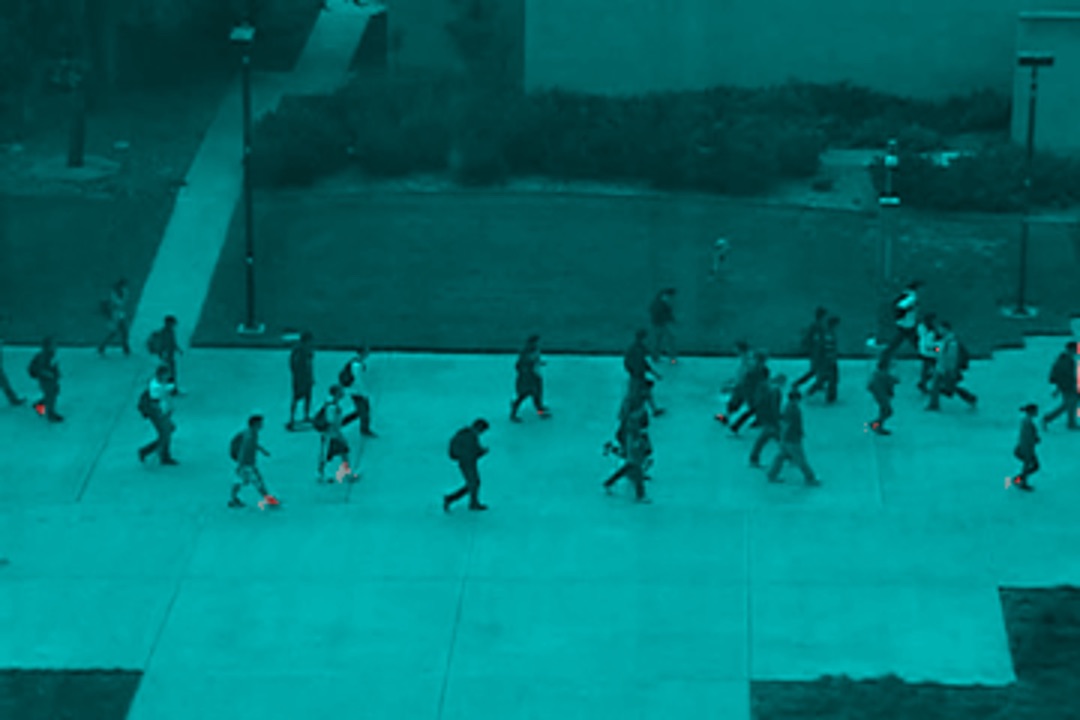}

		\end{center}
		\caption{Sample results of anomaly localization on UCSD: (a) selected from Ped1, (b) Ped2, and (c) confusion cases from Ped1 and Ped2}
		\label{fig:pedvis}
	\end{figure*}
	

	In our experiments we observed that in most of the cases the proposed method hit the abnormality correctly in terms of detection and localization. Only in some cases our measure achieved slightly lower accuracy in anomaly localization and the anomaly detection performance in compare with the state of the art methods. Note that the proposed method is not taking advantage of any kind of learning in comparison with the others. The proposed method can be effectively exploited to detect and localize anomaly with no additional learning costs. Qualitative results on Ped1 and Ped2 are shown in Fig.~\ref{fig:pedvis}. The figure shows we could successfully detect different abnormality sources (like cars, bicycles and skateboards) even in the case in which the object can not be recognized by visual appearance alone (\eg, the skateboard). The last row in Fig.~\ref{fig:pedvis} shows the confusion cases, which not detect the abnormal object (the car) and detect normal as abnormal (the pedestrian). Most of the errors (e.g., miss-detections) are due to the fact that the abnormal object is very small or partially occluded (e.g., the skateboard in the rightmost image) and/or has a ``normal'' motion (i.e., a car moves the same speed of normally moving pedestrians in the scene). 

\subsection{Components Analysis}
\label{sec:abl}
 \noindent\textbf{Analysis of the Importance of two Streams:}
The evaluation of the TCP-only version is performed on ped1 and in two levels: frame-level for anomaly detection, and pixel-level for anomaly localization. In the both cases we unchanged the same experimental setup reviewed in Sec. \ref{sec:exp}.

In the frame-level evaluation, the TCP-only version obtains 93.6\%(AUC), which is slightly lower than 95.7\% of the fused version. In pixel-level evaluation, however, the performance of the TCP-only version dropped 9.3\% with respect to the fused version. This result is still significantly above most of the methods in Tab.1, but this clearly shows the importance of the optical-flow stream for abnormality localization. This is probably due to refining the abnormal segments leveraging the fine motion segments created by the optical-flow map. Hence, fusing appearance and motion can refine the detected area, which leads to a better localization accuracy. Fig. \ref{fig:rocstreams} shows ROCs for the three different states TCP-only, motion-only, and fusion. We simply select equal weights for optical-flow and TCP. 

\noindent\textbf{Binary Quantization Layer vs. Clustering:}
The Binary Quantization Layer ($hconv6$) is a key novelty of our method, which ensures the CNN will work both in the plug-and-play fashion as well as -possibly- being trained end-to-end. In order to evaluate the proposed binary quantization layer, the $hconv6$ removed from the network and a k-means clustering ($k= 2^7$) is performed on the $pool5$ layer of FCN in an offline fashion. Then, the TCP measure is computed on the codebook generated by clustering instead of the binary codes. We evaluated this on UCSD (ped1), obtaining 78.4\% (17.3\% less than our best result on frame-level).

	\section{Discussion}
	
The underlying idea of the proposed approach is to capture the crowd dynamics, by exploiting the temporal variations of CNN features. The CNN network is specifically used to narrow down the semantic gap between low-level pixels and high-level concepts in a crowd scene. The proposed approach provides the means to inject such semantics into model, while maintaining the method complexity in a manageable level. 
	The proposed BFCN is composed of a fully convolutional neural network followed by a binary quantization layer (BQL). The weights of the former network are borrowed from an already pre-trained network and the weights of the BQL layer are obtained through an external hashing module and ``plugged'' into the network as an additional convolutional layer. The last layer of the network (BQL) provides the means to quantize the $pool5$ feature maps into 7-bit binary maps. The training of ITQ is done only once in an off-line fashion and is used for all the experiments without any further fine-tuning. The plug-and-play nature of our proposed architecture enables our method to work across multiple datasets without specific retraining.

	The key idea behind this work is that the consecutive frames should have similar binary patterns, unless they undergo a large semantic change (e.g., abnormal object/motion). The role of TCP measure is to capture such changes across time by computing the irregularity over histogram of binary codes. The proposed TCP measure defines abnormal events as \emph{irregular} events deviated from the normal ones, and the abnormality is measured as the uniformity of the histogram of binary codes. Hence, a flat histogram of binary patterns implies more inconsistency in visual patterns so increases the chance of abnormality. Such particular formulation allows to deal with the context-dependent abnormal events. These characteristics make our method unique in the panorama of the measure-based methods for abnormality detection.
	
	\section{Conclusions}
	In this work, we employed a Fully Convolutional Network as a pre-trained model and plugged an effective binary quantization layer as the final layer to the net. Our method provides both spatial consistency as well as low dimensional semantic embedding. We then introduced a simple yet effective unsupervised measure to capture temporal CNN patterns in video frames. We showed that combining this simple measure with traditional optical-flow provides us with the complementary information of both appearance and motion patterns. The qualitative and quantitative results on the challenging datasets show that our method is comparable to the state-of-the-art methods. As future work, we will study plugging a TCP measure layer and fine-tuning this layer with back-propagation. Moreover, exploring the use of  \cite{icip17_gan, ravanbakhsh2017training} as an alternative to Binary Fully Convolutional Net (BFCN) for end-to-end training of abnormality detection would be a potential direction.

{\small
\bibliographystyle{ieee}
\bibliography{egbib}
}

\end{document}